\documentclass[aps,pre,twocolumn,groupedaddress,showkeys]{revtex4-1}
\usepackage{graphicx}  
\usepackage{natbib} 
\begin{document}

\title{Complexity and universality in the long-range order of words}

\author{Marcelo A. Montemurro}
\email[]{m.montemurro@manchester.ac.uk}
\affiliation{The University of Manchester, Faculty of Life Sciences, Manchester, UK}
\author{Dami\'an H. Zanette}
\affiliation{Centro At\'omico Bariloche e Insituto Balseiro, San Carlos de Bariloche, R\'{\i}o Negro, Argentina}
\begin{abstract}
As is the case of many signals produced by complex systems, language
presents a statistical structure that is balanced between order and disorder. Here we
review and extend recent results from quantitative characterisations of the degree of order in
linguistic sequences that give insights into two relevant aspects of language: the
presence of statistical universals in word ordering, and the link between semantic
information and the statistical linguistic structure. We first analyse a measure of relative entropy that assesses how much the ordering of words contributes to the overall statistical structure of language. This measure presents an almost constant value close to 3.5 bits/word across several linguistic families. Then, we show that a direct application of information theory leads to an entropy measure that can quantify and extract semantic structures from linguistic samples, even without prior knowledge of the underlying language.
\end{abstract}

%\pacs{}

\maketitle

\section{Introduction}
There are only a few known cases of systems that have naturally evolved to encode complex information. For much of Earth's history the chemical language of the genetic code has been the prime example. And although there is evidence that non-human animal species have also developed means of non-trivial communication \cite{Frisch1967,Ouattara2009, Kershenbaum2014}, it was not until the emergence of human language that the last major transition in evolution took place \cite{MaynardSmith1995}. This human faculty evolved as an efficient system capable of transmitting sophisticated messages between different brains, becoming closely linked to our higher mental functions \cite{deacon1997}.

As a carrier of highly complex information, human language must operate under the competing requirements of allowing  high information rate and at the same time being robust under communication errors.  These constraints, of both novelty and redundancy, contribute to shape a statistical structure in linguistic sequences that pose them at a balanced point between order and disorder. Recent advances on the analysis of language with methods and concepts from statistical physics and information theory have disclosed a rich structure at various level of linguistic organisation \cite{Zanette2014}. Here we review and extend recent results on the characterisation of linguistic order by means of novel entropy measures. 

In the first part we discuss a measure of relative entropy that specifically quantifies the degree of order in word patterns. We show that this measure presents a universal value when it is evaluated on language samples belonging to 24 linguistic families. While in their evolutionary history different languages have developed a diverse range of underlying rules and vocabularies, the data suggest that their evolution and diversification were constrained to have an almost constant measure of relative entropy.

In the second part, we use use another entropy measure that is capable of quantifying patterns in word distribution that are closely linked to the semantic role of words. Without any prior linguistic knowledge about the underlying language, we show that it is possible to extract the words that are most closely related to the semantic content of a text and, moreover, disclose semantic relationships between them.  

\section{Universality in the entropy of word ordering}

We may ask the question whether the precise balance between structure and randomness in linguistic sequences depends on features of specific languages, or instead represents some universal aspect of the human language faculty. Some linguists have put forward the hypothesis that even all languages share some basic structural features indicative of cognitive constraints  \cite{Greenberg1963, Chomsky1965,Hawkins1983,Hawkins2004}, while others have challenged the existence of such linguistic universals \cite{Evans2009} or argued that cultural, rather than cognitive traits, are responsible for widespread similarities across some linguistic families \cite{Dunn2011}.  Recently, it has been shown that some patterns in word ordering, like the basic arrangement of subject, verb, and object in sentences, depends on the evolutionary and phylogenetic history of language  \cite{Gell-Mann2011}. Despite the controversy on the presence of linguistic universals at the level of language structure,  quantitative aspects of language presenting universal characteristics have been established. The two best known examples are Zipf's \cite{Zipf1935} and Heap's \cite{Heaps1978} laws in language, which refer to universal features related to word frequencies \cite{Montemurro2001, Montemurro2002b,Zanette2005, Altmann2015}. However, quantitative assessment of universality of word ordering in language are rarer. 

The entropy of a symbolic stochastic source is linked to the predictability of the subsequent outcomes of the sequence when past values are known. A high predictability of future values will entail a low level of surprise in the new symbols, and hence a low entropy. Conversely, a perfectly random sequence will have the highest possible surprise in its symbols, and thus will be characterised by high entropy. Although language sequences are not produced by a stochastic source, it is generally assumed that a large collection of language samples represent an ensemble with enough consistency in its statistical structure to allow the application of the standard formalism of information theory. However, one serious hurdle in computing the entropy of language based on the estimation of block probabilities is the presence of long-range correlations that span from hundreds to thousands of words \cite{Schenkel1993, Ebeling1995, Montemurro2002, Alvarez-Lacalle2006,  Altmann2012}. The sample size that would be needed to estimate the required probabilities grows exponentially with block length, thus quickly rendering insufficient any any available linguistic source.  One way in which this problem can be overcome is through the link between entropy and predictability. Non-parametric estimations of the entropy of language based on guessing games---where subjects have to predict future characters based on the past history of the linguistic sequence---were shown to yield useful results even with moderate sample sizes \cite{Shannon1951, Cover1978}.  Along similar lines, the degree of predictability in a sequence determines how much it could be compressed by a lossless compression method. Highly predictable sequences can be compressed further than more random ones. More rigorously, it can be shown that under the assumptions of stationarity and ergodicity the entropy rate of a stochastic source is a lower bound to the length per symbol of any encoding of it \cite{Cover2006}.  This suggests an approach to estimate the entropy of a symbolic sequence based on the use of efficient lossless compression algorithms.  Many of the practical applications of these ideas are based on the complexity measure \cite{Lempel1976} and  compression algorithms \cite{Ziv1977,Ziv1978} proposed by A. Lempel and J. Ziv, which rely on the estimation of redundancy by matchings between future and past substrings in a symbolic sequence. More recently, methods that estimate the entropy directly by string matching without attempting to compress the symbolic sequence have also been shown to be efficient \cite{Wyner1989, Kontoyiannis1994}. Implementations of these methods have proved to work well for symbolic sequences even in the presence of long range correlations as those found in language \cite{Schurmann1996,Kontoyiannis1998, Puglisi2003,Gao2008}, and without requiring very long sequences in order to converge \cite{Lesne2009}. 

In \cite{Montemurro2011} we carried out an analysis of  of 7,077 texts from 8 languages from 5 linguistic families and one language isolate~\footnote{All but the Old Egyptian and Sumerian text sources were obtained from the Project Gutenberg e-text repository (www.gutenberg.org). The Old Egyptian texts were obtained from the page maintained by Dr Mark-Jan Nederhof at the University of St Andrews (www.cs.st-andrews.ac.uk/mjn/egyptian/texts/) as transliterations from the original hieroglyphs. The Sumerian texts were downloaded from The Electronic Text Corpus of Sumerian Literature (www-etcsl.orient.ox.ac.uk/) and consisted of transliterations of the logo-syllabic symbols.} to assess the contribution to word ordering to the statistical structure of language. The entropy, $H$, was estimated for every text by means of methods derived form compression algorithms and string matching. In order to account for the contribution to linguistic structure that comes only from word frequencies and irrespective of word ordering we also computed the entropy of a randomly shuffled version of the texts, $H_s$. To calculate the entropy of the disordered texts, we first computed the total number of possible arrangements between the words in a given text, as follows: 

\begin{equation}
\Omega=\frac{N!}{\prod_{j=1}^{K}n_j!}\, ,
\end{equation}
where $K$ is the size of the vocabulary and $n_j$ represents the number of instances of the word with index $j=1\ldots K$. Then, the entropy can be estimated {\it\`a la} Boltzmann, as follows:

\begin{equation}
\label{entb}
H_s=\frac{1}{N}\log_2\Omega \,.
\end{equation}

Since the random texts lack any linguistic structure beyond word frequencies, the entropy  will be larger than that of the original sequence, $H$. Therefore, one way to quantify the impact of the ordering of words is by means of a relative entropy measure, defined as  $D_s=H_s-H$. In \cite{Montemurro2011} it is shown that for sufficiently long sequences this quantity  is equivalent to the Kullback-Leibler (KL) divergence between the original and disordered sequences. This measure quantifies the degree of order in a linguistic sequence beyond that contributed by word frequencies alone.

\begin{figure*}[ht]
\includegraphics[width=15cm]{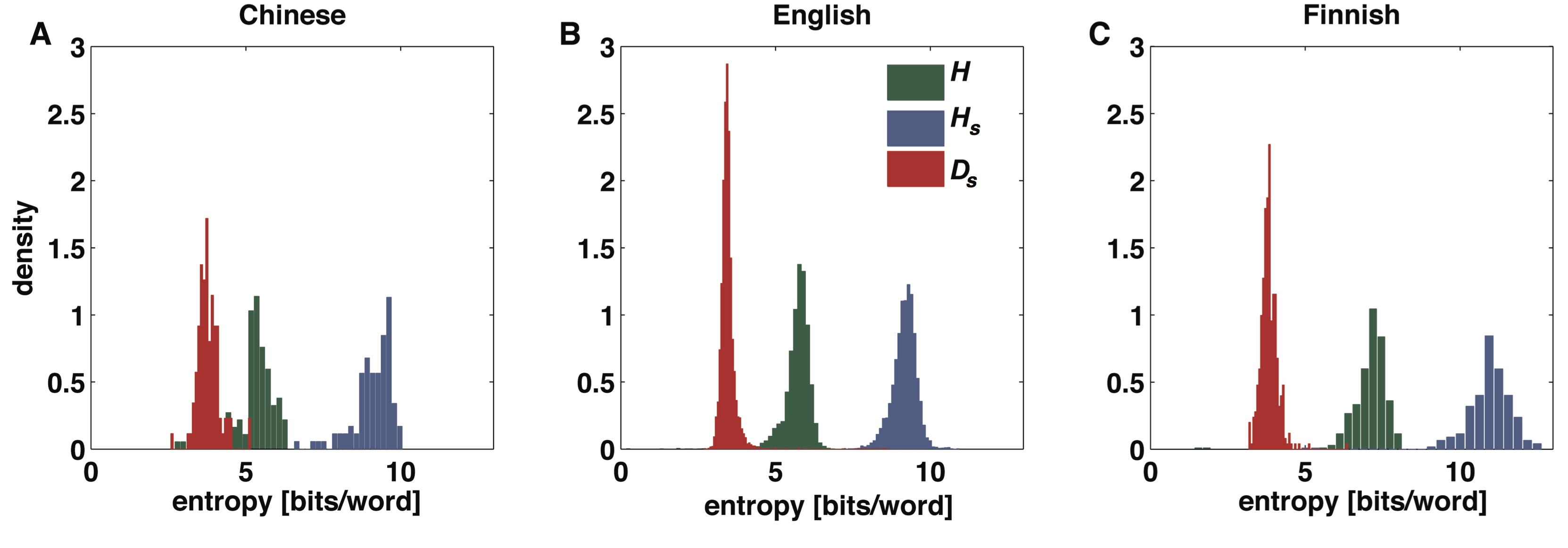}
\caption{{\bf Entropy distributions for corpora belonging to three languages}.
Each panel shows the distribution of the entropy of the random texts lacking linguistic structure (blue); that of the original texts (green); and that of the relative entropy (red). The three languages: Chinese, English, and Finnish, were chosen because they had the largest corpora in three different linguistic families. In panels A, B, and C, the random texts were obtained by randomly shuffling the words in the original ones. In panels D, E, and F, the random texts were generated using the words frequencies in the original texts.. Adapted from \cite{Montemurro2011}.}
\label{fig1}       
\end{figure*}

Figure~\ref{fig1} shows an example of the results obtained for three corpora of languages that differ significantly in their structure: Chinese (Sino-Tibetan), English (Indo-European), and Finnish (Finno-Ugric). In each panel the rightmost distribution corresponds to the entropy of the random texts,  $H_s$, which only accounts for the contribution of word frequencies. While for English and Chinese the value of the distribution of $H_s$ peaks around 9 bits/word, for Finnish it is close to 11 bits/word. The middle distribution for all three panels is that of the entropies of the original texts, $H$. For this quantity there is also language dependence, with English and Chinese having lower values than Finnish.  Finally, the leftmost distribution in each panel is that of  $D_s$. The remarkable feature that emerges from this analysis is that the distribution of $D_s$ peaks at approximately the same value---close to 3.5 bit/word---for the three languages. Furthermore, the distribution of $D_s$ is narrower than that of the direct entropies, suggesting that much of the observed variability in the entropy distributions is due to differences in the vocabulary structure, but overall,  the measure of word ordering given by $D_s$ is less variable over each corpus. 

\begin{figure*}[ht]
\centering
\includegraphics[width=15cm]{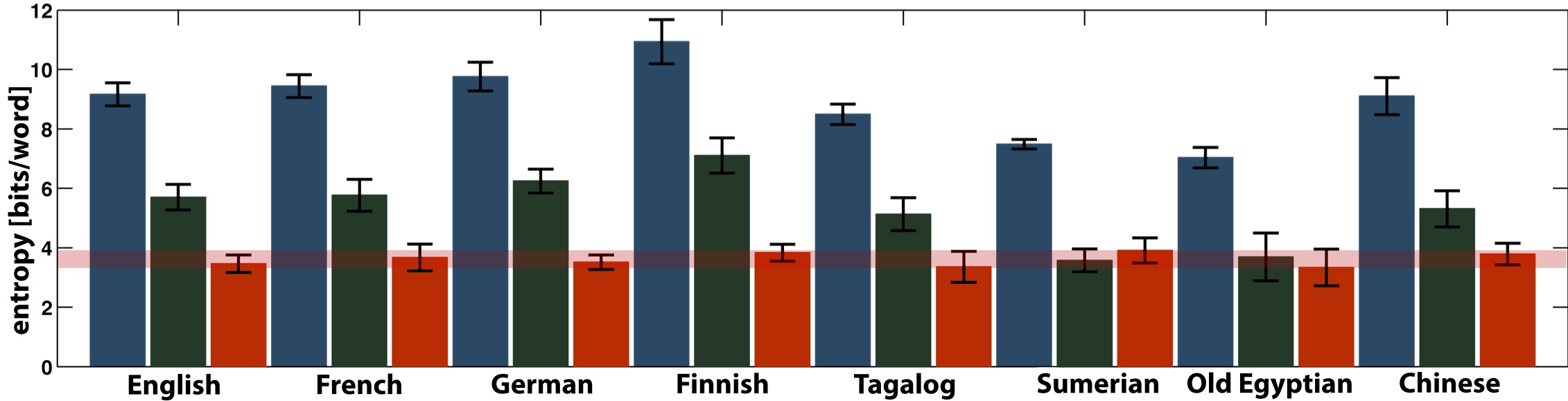}
\caption{{\bf Entropy of eight languages belonging to five linguistic families and a language isolate} (Indo-European: English, French, and German; Finno-Ugric: Finnish; Austronesian: Tagalog; Isolate: Sumerian; Afro-Asiatic: Old Egyptian; Sino-Tibetan: Chinese. The entropies are represented with the same colours as in Figure~\ref{fig1}.Adapted from \cite{Montemurro2011}).}
\label{fig2}       % Give a unique label
\end{figure*}

To verify the generality of these findings we performed a similar calculation for all the 7,077 texts in our 8 corpora. The main results are shown in Figure~\ref{fig2}. Because of the difference in grammar and vocabulary, the values of the two entropies $H$  and $H_s$  show significant variability over the different languages. For example, the entropy of the disordered texts varies from 6.7 bits/word for Old Egyptian to 10.4 bits/word for Finnish, equivalent to a difference of 55\%. Correspondingly, the entropies of the intact texts change from 3.7 bits/word to 7.1 bits/word for the same languages,which is a 91\% difference. However, the relative entropy $D_s$  shows a remarkably consistent value over the different corpora: for Old Egyptian and Finnish the values of $D_s$  are 3.0 and 3.3 bit/word respectively, amounting to only 11\% difference.  We can define the relative variability as the standard deviation of the entropies divided by the mean entropy across languages.  This quantity is 0.14 for $H$, 0.23 for $H_s$, and 0.07 for the relative entropy $D_s$. This fact suggests that while the overall complexity of linguistic structure depends on features specific to each language, a quantification of word ordering given by the relative entropy $D_s$  emerges as a universal feature across languages.

This remarkable constancy of the relative entropy across several linguistic families can be interpreted with the help of simple models of language. In particular, we explored Markovian models of language consisting of just a few words where all quantities of interest could be readily computed \cite{Montemurro2011}. From the analysis of these simple models it turns out that in order to keep the KL divergence constant, an increase in the entropy of the random version of the texts---which is linked to the degree of diversity in the vocabulary--- needs to be accompanied by a corresponding decrease in the range of correlations. Moreover,  it is shown that this patterns can also be found in real languages, suggesting the same explanation for the constancy of the relative entropy  \cite{Montemurro2011}.

\begin{figure}[ht]
%\sidecaption
\centering
\centering
\includegraphics[width=8.5cm]{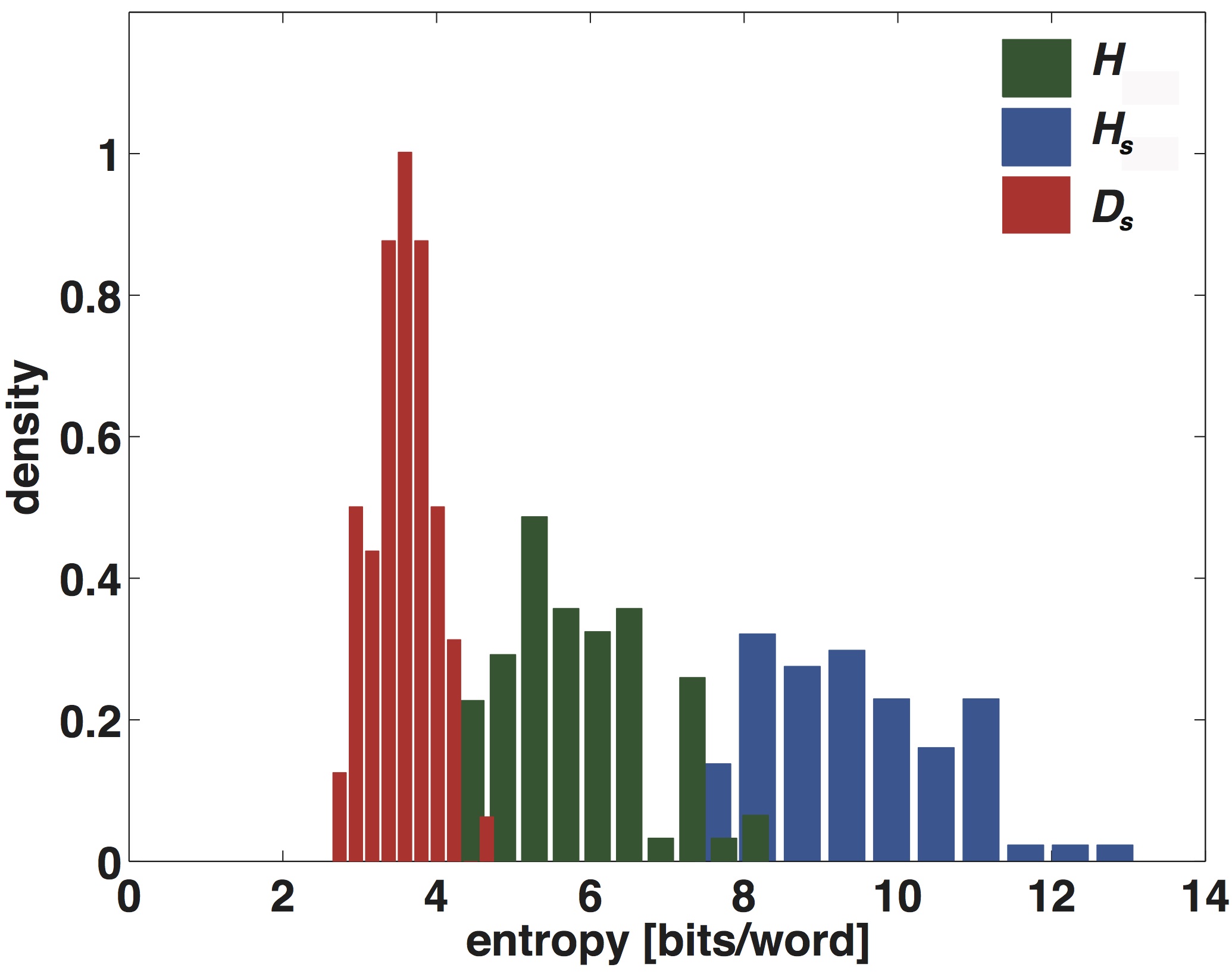}
\caption{{\bf Entropy distributions for a corpus of Bible translations into 75 languages from 24 linguistic families}. The distribution of the entropy of the random texts in shown in blue, that of the original texts in green, and that of the relative entropy in red.}       
\label{fig3}
\end{figure}

Here we extend the previous analysis by presenting results that include several other linguistic families. We analysed a parallel corpus of translations of the Bible into 75 languages from 24 families.\footnote{The parallel Bible corpus was compiled by Christos Christodoulopoulos from the Cognitive Computation Group at the University of Illinois, and downloaded from http://homepages.inf.ed.ac.uk/s0787820/bible/. Only texts that were encoded in Latin characters were used.} Figure~\ref{fig3} shows the distribution of the entropies for all the texts in the corpus pooled together. Consistent with the results shown in Figure~\ref{fig1}, the distribution of the entropy of word ordering $D_s$ is much narrower than that of the direct entropies having a mean value of 3.56 bits/word. Taking the whole corpus, the standard deviation in bits/word of $H$ is 0.94, that of $H_s$ 1.1 and that of $D_s$ 0.4. In Table~\ref{tab1} we show the results grouped according to family, indicating as well the numbers of texts in each group.

\begin{table}
\centering
\caption{Entropy values for the Bible translations grouped into 24 linguistic families}
\begin{tabular}{p{2.5cm}p{1.5cm}p{1.5cm}p{1.5cm}p{1.5cm}}
 \hline\noalign{\smallskip}
 Family & Languages & $H_{s}$ [bits] & $H$ [bits] & $D_s$ [bits]\\
%\noalign{\smallskip}\svhline\noalign{\smallskip}
Afro-Asiatic & 5 & 9.51 & 5.83 & 3.69 \\
 Algic & 1 & 9.68 & 6.48 & 3.20 \\
 Altaic & 1 & 10.85 & 6.52 & 4.33 \\
 Arawakan & 1 & 10.97 & 7.60 & 3.36 \\
 Austro-Asiatic & 1 & 9.09 & 5.43 & 3.66 \\
 Austronesian & 7 & 8.67 & 5.32 & 3.35 \\
 Basque & 1 & 10.89 & 7.23 & 3.65 \\
 Caribean & 1 & 7.37 & 4.39 & 2.98 \\
 Chibchan & 1 & 8.23 & 4.97 & 3.26 \\
 Esperanto & 1 & 9.28 & 5.72 & 3.55 \\
 Creole$^b$ & 2 & 7.75 & 4.88 & 2.88 \\
 Equatorial & 1 & 9.73 & 5.79 & 3.94 \\
 Indo-European & 25 & 9.65 & 6.05 & 3.60 \\
 Jivaroan & 3 & 10.99 & 7.13 & 3.84 \\
 Mayan & 6 & 8.44 & 4.67 & 3.76 \\
 Niger-Congo & 6 & 10.19 & 6.58 & 3.62 \\
 Nilo-Saharan & 2 & 8.14 & 5.23 & 2.92 \\
 Oto-Manguean & 2 & 7.57 & 4.40 & 3.17 \\
 Quechuan & 1 & 10.92 & 7.41 & 3.51 \\
 Sino-Tibetan & 1 & 9.37 & 6.05 & 3.32 \\
 Tucanoan & 1 & 8.39 & 4.38 & 4.00 \\
 Finno-Ugric & 3 & 10.78 & 6.83 & 3.95 \\
 Uto-Aztecan & 1 & 9.27 & 5.97 & 3.30 \\
 West Papuan & 1 & 8.22 & 4.83 & 3.38 \\
% \noalign{\smallskip}\hline\noalign{\smallskip}
\end{tabular}
\\
\flushleft{ $^a$ The two creole languages are Aukan and Haitian }
\label{tab1}
\end{table}

In this first section we reviewed and extended evidence for the universal value of the relative entropy across human language.  In the next section, we will discuss another relative entropy measure that, by assessing the specificity words to different contexts, can quantify and extract semantic information from language samples for which essentially no prior knowledge of the underlying linguistic structure is required.

\section{An entropy measure of semantic information}

In linguistic sequences both grammatical and semantic constraints affect the specific use of words at different ranges.  At scales of the typical sentence length both grammatical and semantic constraints affect linguistic structure, whereas order at longer scales is mostly shaped  by semantic requirements. Consequently, words that are specific to the topics being addressed in a text show a different overall distribution compared to words that have a more structural role in language. In particular, several studies have confirmed that the words more relevant to the topics in a text tend to have an irregular distribution characterised by clustered, or bursty,  patterns of occurrence \cite{ Harter1975, Church1995, Ortuno2002, Herrera2008, Altmann2009, Montemurro2010, Carretero2013}. On the contrary, function words, which are no context-specific, appear more uniformly distributed. Along similar lines we have previously reported a measure based on the entropy in the distribution of words over a text that could be used to discriminate between words belonging to different grammatical classes \cite{Montemurro2002a} 

This insight can be incorporated into a measure of semantic information based on information theory.  A measure of the information in the word distribution is based on the observation that the relevant words, or keywords, in a text are typically more dependent on the specific thematic context than non-informative words. Therefore, the specific distribution of these words can be used to distinguish statistically different parts of a text.  For example, a word that only appears in one specific chapter of a book is a perfect tag for that chapter, i.e. if that word is found, it is known with certainty which chapter is being read. Despite the majority of words will have a less concentrated distribution over the text,  the non-uniformities in their distribution can still be used to link them to specific contextual domains.

Consider a text of $N$  words in length, containing $K$  different words. The text is divided into $P$ equal parts, of length $s=N/P$. For every word $w$  that appears $n$ times in the text, we can define its distribution over the text as the probability $p(w|j)$ of finding that particular word in part  $j$ $(j=1\ldots P )$. This probability is estimated as the ratio $n_j/s$, where $n_j$  is the number of occurrences of word $w$  in part $j$, and is normalised as $\sum_{w=1}^Kp(w|j)=1$.   Let us call $p(j)=1/P$  the {\em a priori} probability that a given word $w$ appears in part $j$, then the overall probability of occurrence of the word is $\sum_{j=1}^P p(w|j)p(j)=p(w)$, where $p(w)=n/N$. After observing an instance of word $w$, the probability that it comes from part $j$  is given by $p(j|w)$, which can be computed as $p(j|w)=p(w|j)p(j)/p(w)$, or explicitly in terms of word occurrences as  $p(j|w)=n_j/n$. Then, the mutual information between the sections of the text and the distribution of words is \cite{Cover2006}: 
 			
\begin{equation}
\label{minfo}
M(J,W)=\sum_{w=1}^K p(w) \sum_{j=1}^P p(j|w) \log_2 \frac{p(j|w)}{p(j)}\,.
\end{equation}

Words that appear in the text a number of times $n\ll N$ will have statistical fluctuations in their distribution over the partition that may induce an overestimation of the mutual information. We can correct for this bias by subtracting the mutual information computed over randomised versions of the text obtained by shuffling all the words positions. Despite it being computed over random versions of the text---where all the relationships between the words and its original contexts is lost---this quantity will not be zero in general due to the presence of statistical fluctuations. Let us call  $\hat{M}(J,W)$ the mutual information estimated from one realisation of the shuffled text. Then, we can define the information in the distribution of words as $\Delta I(s)=M(J,W)-\langle {\hat M}(J,W) \rangle$, where the average is taken over an infinite number of realisations of the word shuffling. Then, using Eq.~(\ref{minfo}) and regrouping terms leads directly to the following equation for the corrected information:

\begin{equation}
\label{deltai}
\Delta I(s)=\sum_{w=1}^K p(w) \left[\langle {\hat H}(J|w)\rangle-H(J|w)\right]\, ,
\end{equation}

\noindent where the sum is taken over the whole vocabulary of  $K$  words.  Thus, Eq.~(\ref{deltai}) represents an information measure quantifying the degree of specificity of words over contextual domains characterised by the scale $s$. The entropy term  $H(J|w)$  can be directly computed from the word counts across the  $P$  parts of the text as follows:

\begin{equation}
\label{entropy}
H(J|w)=-\sum_{j=1}^P \frac{n_j}{n} \log_2 \frac{n_j}{n} \, .
\end{equation}

This quantity indicates how non-uniform is the use of word $w$ over the text: the smaller the entropy the more non-uniform its distribution. The other entropy that appears in Eq.~(\ref{deltai}), $\langle {\hat H}(J|w) \rangle$, is similar to $H(J|w)$ with the difference that it is computed over randomly shuffled versions of the text, and averaged over an infinite number of realisations of the shuffling. This last term accounts for the fluctuations that are expected due to the finite number of occurrences of words and can be computed analytically \cite{Montemurro2010}. The information given by $\Delta I$ is then a an average measure of how non-random is the distribution of words over a text.

We can gain insight into the meaning of Eq.~(\ref{deltai}) by analysing the behaviour of the two entropy quantities for an actual text. Figure~\ref{fig3} shows the entropies as functions of frequency computed for all the words in {\it The Analysis of Mind} by Bertrand Russell. Yellow dots show the entropy obtained after a random shuffling of all the words in the text and the black line is the average taken over an infinite number of random shufflings as given by the analytical estimation of $\langle {\hat H}(J|w) \rangle$ (see \cite{Montemurro2010} for details). The entropy of the randomly distributed words versus frequency shows the overall trend that is expected as a consequence of statistical fluctuations in the ordering of words. High frequency words in the random  text will typically have more uniform distributions than low frequency words.  Finally, the entropy for the words in the original text are represented as blue dots. While the same overall trend as for the random text is observed, most of the words in the original text show values of the entropy significantly smaller than those in the randomly shuffled text for the same frequency. This is a consequence of the strong ordering constraints imposed by the linguistic structure in the original texts.   Therefore, the contribution of a word to the total information is equal to this entropy difference weighted by the frequency of the word.

\begin{figure}[ht]
\centering
\includegraphics[width=9.5cm]{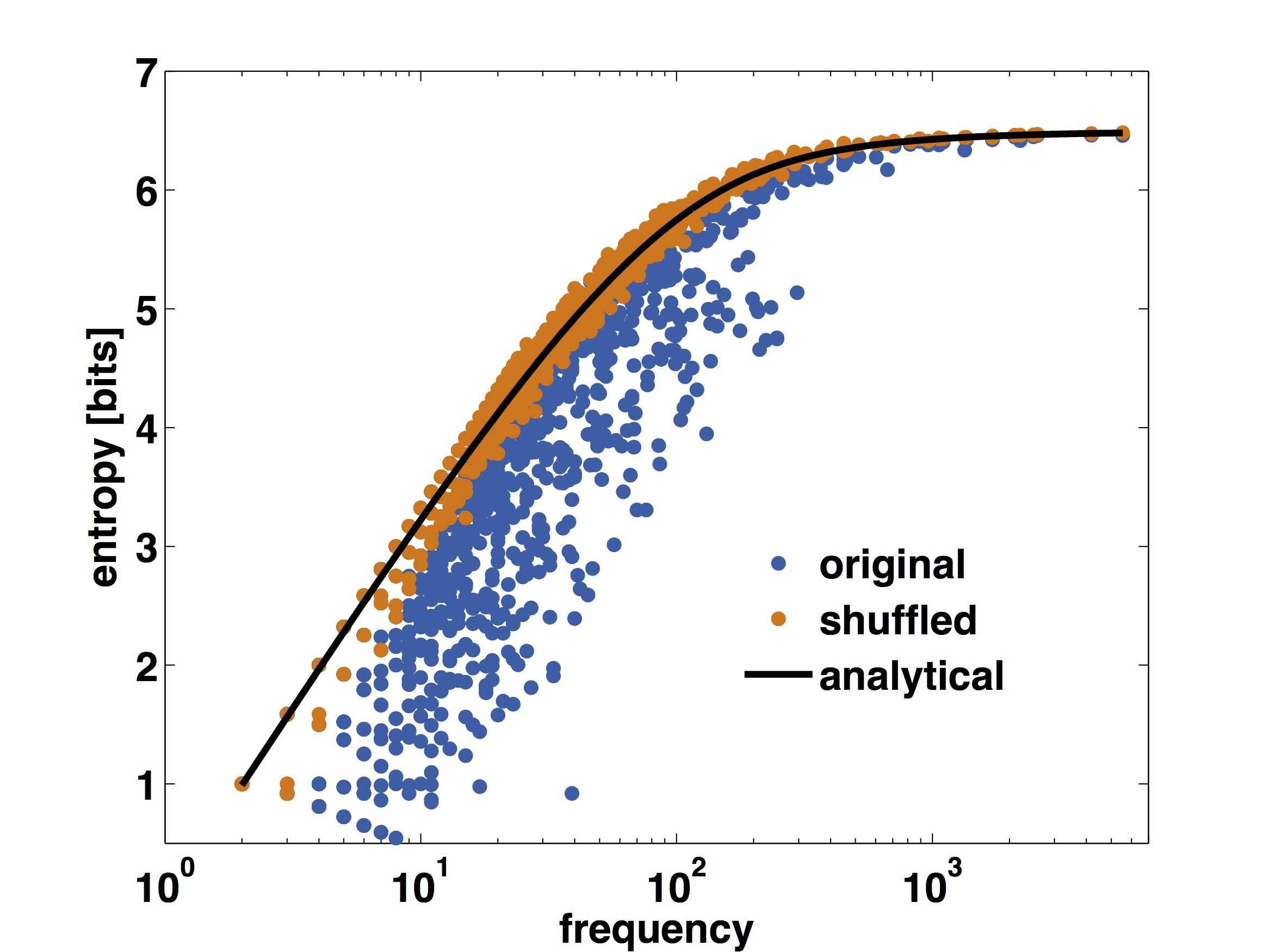}
\caption{
{\bf Entropy of words in real and random texts}. Black dots correspond to the entropy $H(J,w)$ of words in {\em The Analysis of Mind} (see Eq.~(\ref{entropy})),  using the scale at which the information in the distribution of words is maximal (s=750 words); grey dots represent the entropies computed over a randomly shuffled version of the text. The black full line corresponds to the analytical estimation of entropy of the random text averaged over an infinite number of the realisations of the shuffling (see \cite{Montemurro2010} for details).}
\label{fig4}    
\end{figure}

 \begin{figure}[ht!]
\centering
\includegraphics[width=9.5cm]{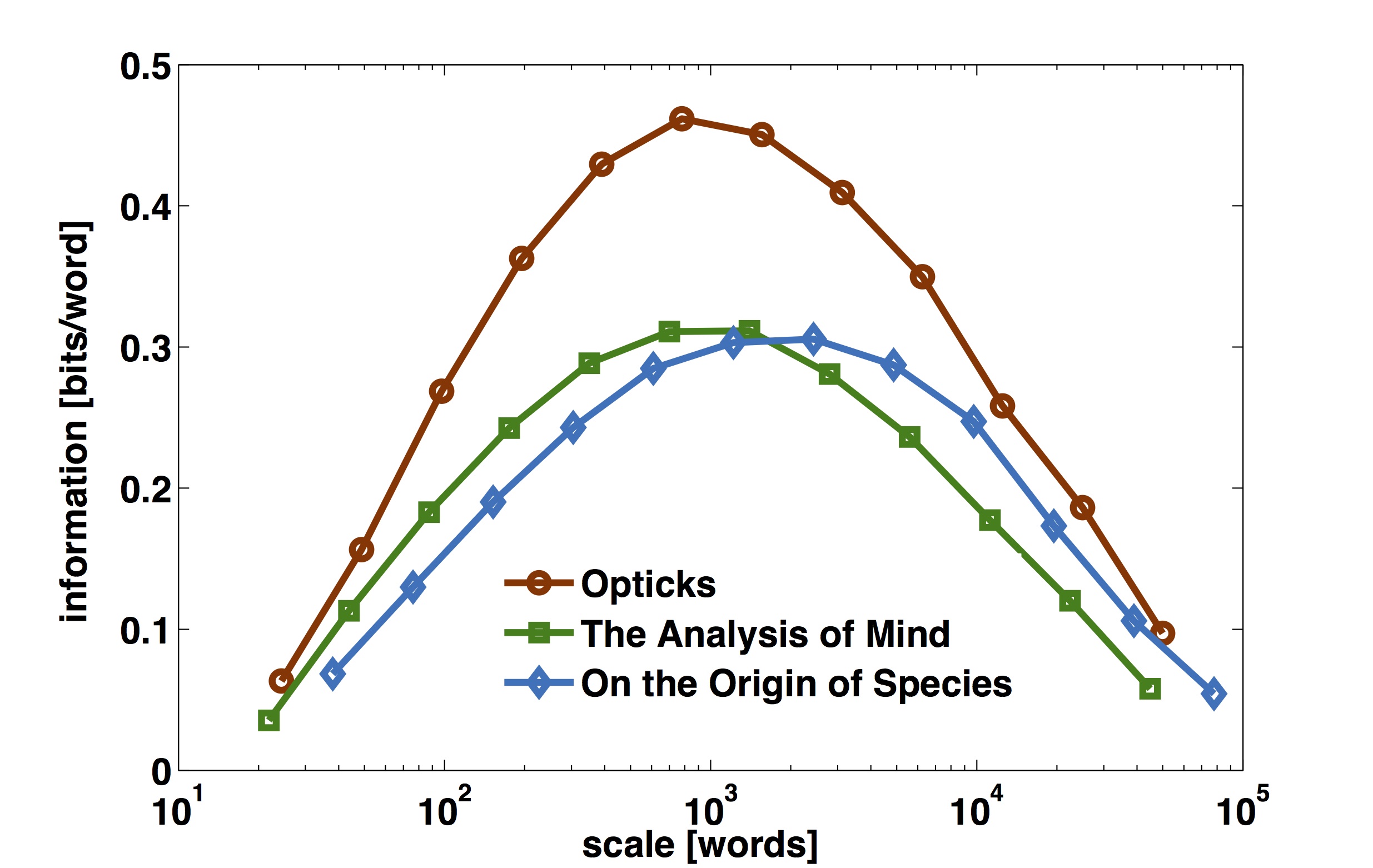}
\caption{
{\bf Information in the distribution of words for three books in English}. The curves represent the estimation of the information given by Eq.~(\ref{deltai}). The texts are {\em Opticks} by Isaac Newton, {\em The Analysis of Mind}, by Bertrand Russell, and {\em On the Origin of Species}, by Charles Darwin. For each text the information in bits/word is shown as a function of the scale parameter determining the size of contextual domains.}
\label{fig5}      
\end{figure}

Figure~\ref{fig5} shows the information in the distribution of words for three books in English: {\em Opticks} by Isaac Newton, {\em The Analysis of Mind}, by Bertrand Russell, and {\em On the Origin of Species}, by Charles Darwin. For small and large values of the scale parameter setting the size of the contextual domains, the value of the information approaches zero. This is because in the extremes the distribution of words in the real and random texts become identical over the partition. At intermediate values all distributions show an optimal value of the scale at which the difference in the distribution of words between the original and shuffled texts is maximised. This optimal scale represents the typical size of the partition at which the distribution of words over the text becomes most heterogeneous.  For the three texts shown in Figure~\ref{fig5} the values of $s$---in number of words---at which the information is maximal is close to 950 for {\em Opticks}, 750 for {\em The Analysis of Mind}, and 1930 for {\em On the Origin of Species}.  The range of the optimal scales is much larger than the scope of grammatical rules and is determined by the semantic structure of the texts. An analysis done on more than 5,000 books written in English supports the conclusion that the optimal scale is related to the typical size of semantic domains over which subtopics are developed \cite{Montemurro2010}.

\begin{table}[ht!]
\centering
\caption{List of most informative words for three texts}
\begin{tabular}{p{2.0cm}p{2.5cm}p{3cm}}
 \hline\noalign{\smallskip}
{\em Opticks} & {\em Origin of Species} & {\em The Analysis of Mind}\\
%\noalign{\smallskip}\svhline\noalign{\smallskip}
     rings   &     on   &     image   \\
     colours   &     species   &     memory   \\
     prism   &     varieties   &     images   \\
     paper   &     hybrids   &     word   \\
     the   &     forms   &     belief   \\
     red   &     islands   &     words   \\
     light   &     of   &     desire   \\
     I   &     will   &     sensations   \\
     rays   &     selection   &     you   \\
     glass   &     genera   &     past   \\
     bodies   &     plants   &     knowledge   \\
     colour   &     seeds   &     box   \\
     image   &     sterility   &     content   \\
     was   &     fertility   &     consciousness   \\
     blue   &     characters   &     appearances   \\
     refraction   &     breeds   &     movements   \\
     water   &     groups   &     mnemic   \\
     greek   &     water   &     feelings   \\
     lens   &     the   &     proposition   \\
 %\noalign{\smallskip}\hline\noalign{\smallskip}
\end{tabular}
\label{tab2} 
\end{table}

\subsection{Information in the distribution of individual words}

From Eq.~(\ref{deltai})  it is apparent that the total information is a sum of contributions from individual words. Each word can then be assigned an information value equal to its weight in the sum, as $\Delta I_w(s)=p(w)\left[\langle {\hat H}(J|w)\rangle-H(J|w)\right]$. This means that the information associated with individual words depends both on their frequency and on the difference of the entropies computed on the real text and on a random version of it. In order to be informative, a word must be frequent and at the same time have a heterogeneous distribution over the text as a consequence of its specificity to its relevant semantic contexts. When words are ranked by their contribution to the overall information in a text, the top words are those more closely related to its semantic content  \cite{Montemurro2010}. Table~\ref{tab1} lists the most informative words for the three books used in Figure~\ref{fig5}, where the information was estimated at the optimal scale in each case. The majority of these words  relate closely to the major themes in each book. The few cases of functional words that appear in the list are due to the fact that fluctuations in their distribution are greatly magnified by their high frequency. It is interesting to note that the only knowledge about the structure of the language that is incorporated into the calculation of the information is the distinction between word tokens.

\subsection{Extraction of semantic networks}  

\begin{figure*}[ht!]
\centering
\includegraphics[width=15cm]{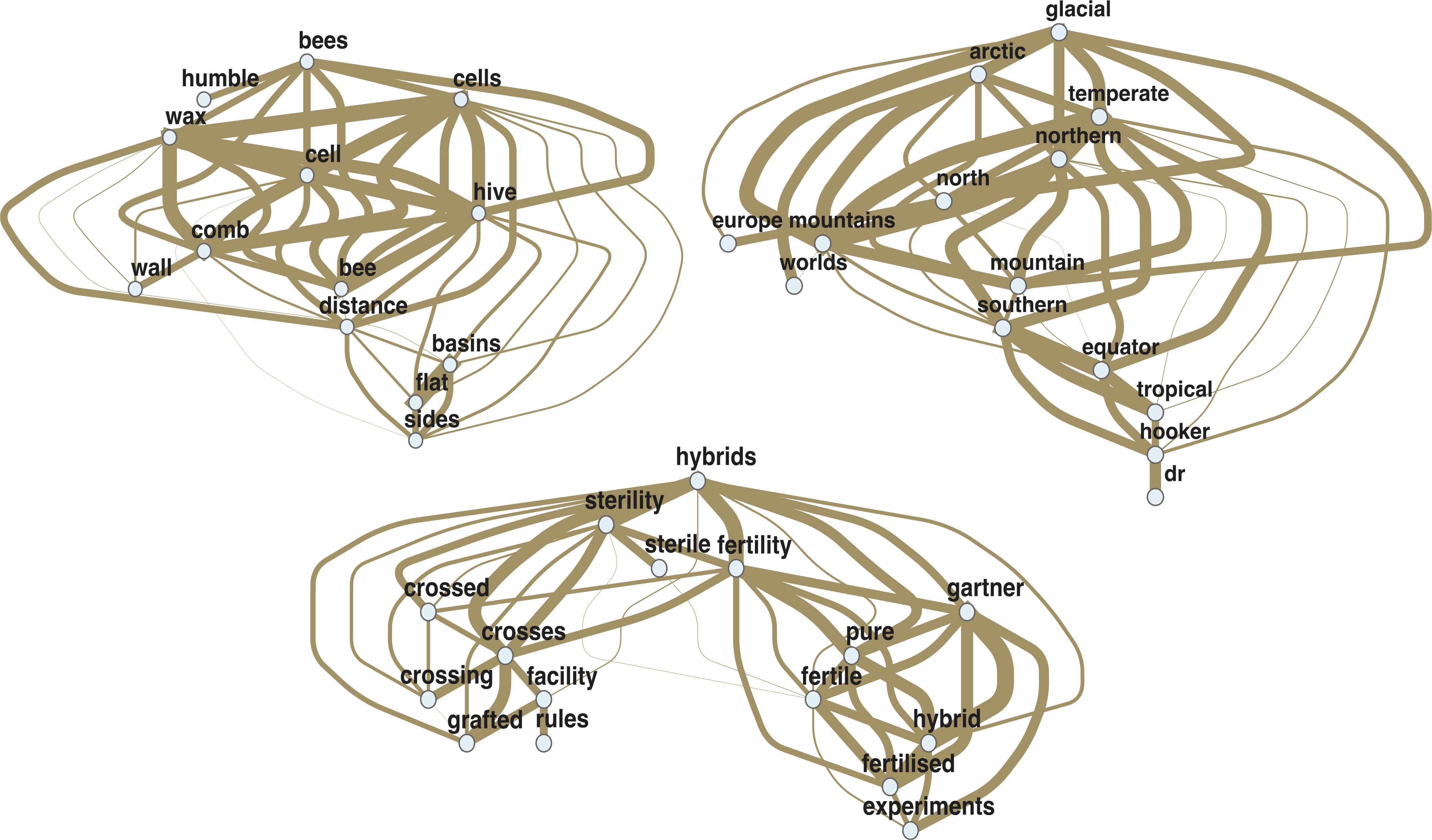}
\caption{
{\bf Semantic networks from {\em On the Origin of Species}}. The networks are examples obtained form the analysis of co-occurrences of the 500 most informative words without any prior knowledge about the underlying linguistic structure. The thickness of the edges indicate the strength of the connections.}
\label{fig6}      
\end{figure*}

Once the most informative words are identified, it is possible to study their co-occurrences in a systematic way with the aim of identifying  groups of related words that tend to appear in similar contexts. Words for which their co-occurrence is statistically robust will most likely present semantic relationships. Both in \cite{Montemurro2013} and \cite{Montemurro2014} the information-based keyword extraction was complemented with a word-space analysis in order to capture the structure of semantic networks. As an example, Figure~\ref{fig6} shows three semantic networks obtained from {\it On the Origin of Species} by Charles Darwin, where words in each network clearly relate to each other.
 
\section{Conclusions}
We have presented a summary of recent progress on the characterisation of the structure of language by means of entropy measures. We first addressed the controversial question of whether there are universal statistical patterns in word ordering.   We showed that while estimations of the direct entropy over different languages yield values that strongly depend on the particular language, a measure of relative entropy that specifically quantifies the degree of word ordering presents an almost constant value over a wide range of linguistic families. This relative entropy measure can be shown to be equivalent to the Kullback-Leibler divergence between a linguistic sequence and a disordered version of it. We have also discussed some steps towards the interpretation and implications of this constancy. By using simple models and analysis of real languages, we showed that the constancy of the relative entropy requires an interplay between the diversity of their vocabularies and the extent of correlations. The degree of universality shown in this feature suggests that languages from a wide range of families have evolved under the precise constraint of keeping the relative entropy constant. 

In the second part of this communication we discussed an information measure that quantifies how informative is the distribution of words in a text over the different sections of a partition of it. This quantity also relies on the assessment of the balance between order and disorder in language. One interesting insight provided by this measure is the presence of a scale in language which seems to be related to the typical lengths---in words---over which specific topics are developed in language. Moreover, since the information is additive over the words in a text    it is possible to ascribe an individual information value to each word defined as its weight in the overall sum, which  allows the ranking of words by their frequency distribution. The words that contribute the most to the information turn out to be the ones most closely related to the semantic content of the texts, thus providing an method for automatic keyword extraction that requires essentially no knowledge of the underlying linguistic structure of the texts.

The methods described in this paper show that a careful assessment of the balance between order and disorder in linguistic sequences with methods from statistical physics and information theory can offer significant clues into the structure of language. Still, many questions remain open, as for example what are the actual universal mechanisms that constrain the evolution of language along trajectories of constant Kullback-Leibler divergence, or whether there are further insights into the link between {\em meaning} and statistics.

% Create the reference section using BibTeX:
\bibliography{language_bib}

\end{document}